%% file: main.tex
\documentclass[letterpaper, 10 pt, conference]{ieeeconf}  

\IEEEoverridecommandlockouts                              

\usepackage{graphics} 
\usepackage{graphicx}
\usepackage{amsmath} 
\usepackage{amsfonts}
\usepackage{xcolor}
\usepackage{wrapfig}
\usepackage{booktabs}

\title{\LARGE \bf
Sample Efficient Optimization for Learning Controllers \\ for Bipedal Locomotion
}

\author{Rika Antonova$^{*,1}$, Akshara Rai$^{*,1,2}$ and Christopher G. Atkeson$^{1}$
\thanks {* Both of these authors contributed equally.}
\thanks{$^{1}$ Robotics Institute, School of Computer Science, Carnegie Mellon University, USA.}
\thanks{$^{2}$ Autonomous Motion Department, MPI-IS, T\"{u}bingen, Germany.}
\thanks{\tt\small \{rantonov, arai\}@andrew.cmu.edu, cga@cs.cmu.edu}
\thanks{This research was supported in part by the Max-Planck-Society. Any opinions, findings, and conclusions or recommendations expressed in this material are those of the author(s) and do not necessarily reflect the views of the funding organization.}
}

\begin{document}

\maketitle
\thispagestyle{empty}
\pagestyle{empty}

\begin{abstract}
Learning policies for bipedal locomotion can be difficult, as experiments are expensive and simulation does not usually transfer well to hardware. To counter this, we need algorithms that are sample efficient and inherently safe. Bayesian Optimization is a powerful sample-efficient tool for optimizing non-convex black-box functions. However, its performance can degrade in higher dimensions. We develop a distance metric for bipedal locomotion that enhances the sample-efficiency of Bayesian Optimization and use it to train a 16 dimensional neuromuscular model for planar walking. This distance metric reflects some basic gait features of healthy walking and helps us quickly eliminate a majority of unstable controllers. With our approach we can learn policies for walking in less than 100 trials for a range of challenging settings. In simulation, we show results on two different costs and on various terrains including rough ground and ramps, sloping upwards and downwards. We also perturb our models with unknown inertial disturbances analogous with differences between simulation and hardware.
These results are promising, as they indicate that this method can potentially be used to learn control policies on hardware.
\end{abstract}


\section{INTRODUCTION}
\input{sec_intro}

\section{BACKGROUND}
\input{subsec_background_bo}

\input{subsec_background_locomotion}

\input{subsec_neuromuscular_models_cmaes}

\section{Determinants of Gait Kernel}
\input{subsec_bo_kernel}

\input{subsec_new_distance_metric}

\section{Experiments}
\input{subsec_experimental_setup}

\input{subsec_model_disturbances}

\input{subsec_experiments_kernel}

\section{CONCLUSIONS}
\input{sec_conclusions}

\bibliographystyle{plain}
\bibliography{main}

\end{document}

%% file: sec_intro.tex
Designing and learning policies for bipedal locomotion is a challenging problem, as it is extremely expensive to do such experiments on an actual robot. We typically do not have robots that can take a fall, and it is cumbersome to perform these experiments. On top of this, most objective functions are non-convex, non-differentiable and noisy. With these considerations in mind, it is important to find optimization methods that are sample efficient, robust to noise and non-convexity, and try to minimize the number of bad policies sampled. Bayesian Optimization is one such gradient-free black-box global optimization method, that is sample efficient and robust to noise. 

One common way of learning controllers is to come up with a control policy parameterizations and a cost function \cite{calandra2014bayesian}, \cite{simbicon}, \cite{coros2010generalized}. This cost is now minimised with respect to the policy parameters. In general, a variety of optimization approaches could be applied to this problem, for example gradient descent, evolutionary algorithms, and random search. Approaches like grid search, pure random search, and various evolutionary algorithms usually make the least restrictive assumptions, but are not sample-efficient. Gradient-based algorithms can be very effective when an analytical gradient is available or can be approximated effectively. However, random restarts are usually necessary to optimize non-convex functions to avoid bad local optima, reducing sample efficiency. Bayesian optimization is well suited for such optimization problems as it constructs a global representation of the cost function, while only reducing uncertainty in promising regions of the search space.

In this work, we learn optimal reflex parameters for neuromuscular models described in~\cite{geyer2010muscle}. These models are human-inspired control pathways that are capable of producing locomotion behaviour in a variety of scenarios as demonstrated in ~\cite{song2015neural}. We start with a 2 dimensional \mbox{7-link} simulated robot with hip, knee and ankle actuation. We formulate a cost function which incorporates components like distance and time walked and optimize it over a set of 16 parameters of the neuromuscular model.

Promising results were reported in prior work for optimizing an \mbox{8-dimensional} control policy for a small bipedal robot using Bayesian Optimization~\cite{calandra2014bayesian}. However, our \mbox{16-dimensional} search space proved to be challenging for standard Bayesian Optimization, with performance not much better than uniform random search. Bayesian Optimization can take advantage of a domain-specific kernel, which gives an informed similarity between different policies. In this paper, we achieve this by developing a Determinants of Gait (DoG) kernel for the domain of bipedal ocomotion. Our kernel uses gait characteristics described in~\cite{Saunders} to create an appropriate similarity metric between different parameter sets of the neuromuscular model. Under this metric, policies that generate successful walking gaits are closer together and policies that result in a fall are more distant from successful ones. This helps the optimization to effectively separate good regions of the parameter space from bad regions, making it more sample efficient.

We  demonstrate that our kernel can substantially reduce the number of function evaluations (or robot trials) needed for optimization on different terrains with modeling disturbances. This leads us to believe that our kernel helps improve sample efficiency in conditions different from those in which it was generated. Potentially, we can generate this kernel in simulation and use it for optimization on the actual robots.

%% file: subsec_background_bo.tex
\subsection{Overview of Bayesian Optimization}
\label{subsec_bo_overview}

Bayesian Optimization is a framework for sequential global search to find a vector $\pmb{x}^*$ that minimizes a cost function $f(\pmb{x})$, while evaluating $f$ as few times as possible (\cite{BOtutorial2010} give an overview).
\vspace{-5px}
\begin{align*}
\pmb{x}^* = \arg \min_{\pmb{x}} f(\pmb{x})
\vspace{-5px}
\end{align*}
 
\begin{figure}[t]
\centering
\caption{Posterior and acquisition function in Bayesian Optimization.}
\label{fig_bo_illustration}
\includegraphics[width=0.48\textwidth]{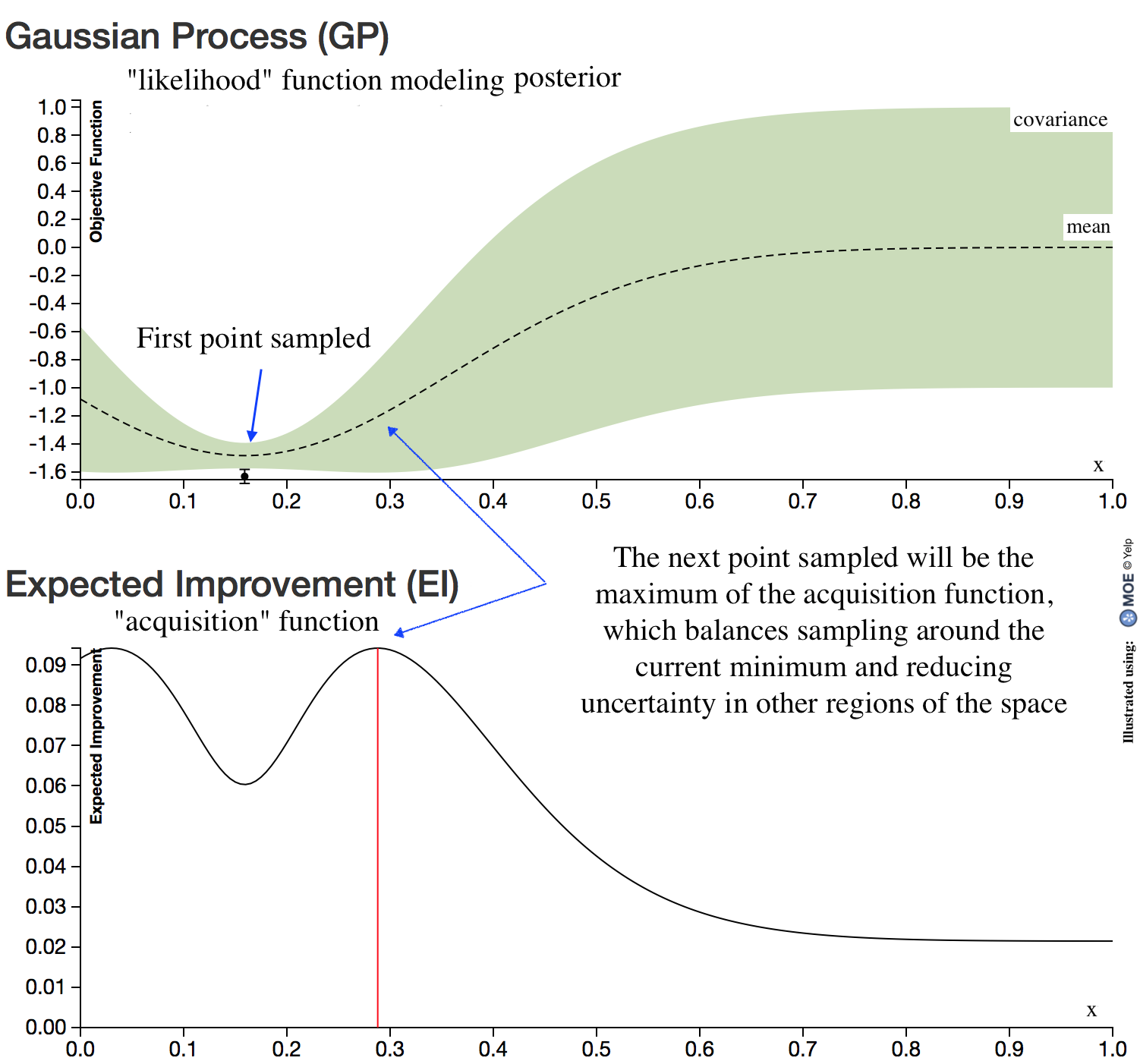}
\vspace{-20px}
\end{figure}

The optimization starts with initializing a prior to capture uncertainty over the value of $f(\pmb{x})$ for each $\pmb{x}$ in the domain. At iteration $t$ an auxiliary function $u$, called an acquisition function, is used to sequentially select the next parameter vector to test, $\pmb{x}_t$. $f(\pmb{x}_t)$ is then evaluated by doing an experiment, and used to update our estimate of $f$.
The aim of the acquisition function is to achieve an effective tradeoff between exploration and exploitation.
The use of an acquisition function is illustrated in Figure~\ref{fig_bo_illustration} (1D example).

A common way to model the prior and posterior for $f$ is by using a Gaussian Process $f(\pmb{x}) \sim \mathcal{GP}(\mu(\pmb{x}), k(\pmb{x}_i, \pmb{x}_j))$, with mean function $\mu$ and kernel $k$. The mean of the prior can be set to $0$ if no relevant domain-specific information is available. The kernel $k(\pmb{x}_i, \pmb{x}_j)$ encodes how similar $f$ is expected to be for two inputs $\pmb{x}_i, \pmb{x}_j$: points close together are expected to influence each other strongly,
while points far apart would have almost no influence. The most widely used kernel is Squared Exponential kernel of the form $k_{SE}(\pmb{x}_i, \pmb{x}_j) = \exp\Big(- \frac{1}{2} \|\pmb{x}_i - \pmb{x}_j\|^2 \Big)$.

\begin{figure}[t]
\centering
\caption{Bayesian Optimization posterior for an example function.}
\label{fig_bo_illustration_posterior}
\includegraphics[width=0.48\textwidth]{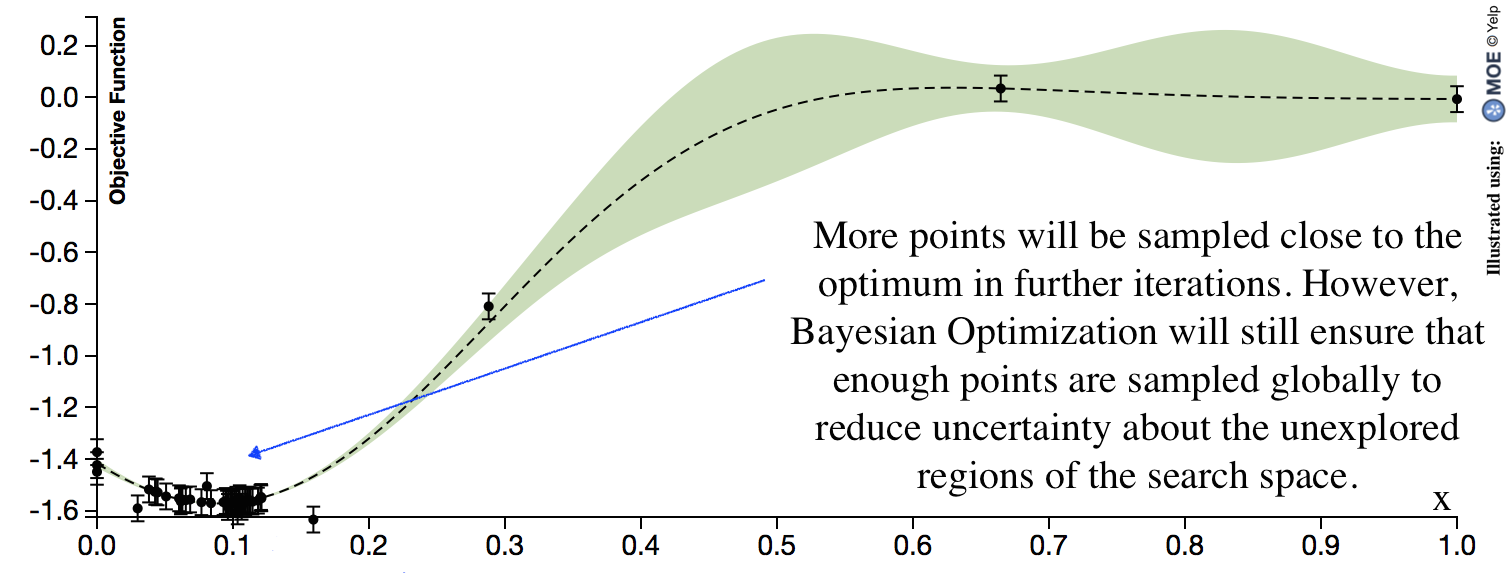}
\vspace{-20px}
\end{figure}

A Gaussian Process conditioned on cost evaluations represents a posterior distribution for $f$. 
Update equations for the posterior mean and covariance conditioned on evaluations can be found in~\cite{GPsMLBook} for both noisy and noiseless settings. An example posterior is illustrated in Figure~\ref{fig_bo_illustration_posterior}.

%% file: subsec_background_locomotion.tex
\subsection{Optimization for Bipedal Locomotion}
\label{subsec_background_locomotion}

Bayesian Optimization (BO) with Gaussian Processes and closely related methods have been recently applied to several robotics domains. Krause \textit{et al.}~\cite{krause2008} developed an approach utilizing Gaussian Processes and the principle of optimizing mutual information for solving sensor placement problems. Martinez-Cantin \textit{et al.}~\cite{mantinez2009} used BO for online path planning for optimal sensing with a mobile robot. Lizotte \textit{et al.} \cite{lizotte2007} used a closely related approach of Gaussian Process Regression to optimize the gait on a quadruped robot and showed that this approach required substantially fewer evaluations than state-of-the-art local gradient approaches.

More specific to the domain of bipedal locomotion, Calandra \textit{et al.} used BO to efficiently find gait parameters that optimize a desired cost~\cite{calandra2014bayesian}. They optimized eight parameters - four threshold values of a finite state machine of a walking controller and four control signals applied during extension and flexion of knees and hips - for a small biped. 

While these previous results are encouraging, it is not immediately clear whether BO would be as successful in finding good policies for higher-dimensional controllers. Calandra \textit{et al.} mentioned that only around 1\% of the parameter space they considered led to walking gaits, and we have observed similar difficulties in our experiments in 16 dimensions. Hence two questions arise : would BO be effective if the dimensionality is increased from 8 to 16? And if it does, how does it compare to previously used approaches, like CMA-ES~\cite{song2015neural}?

%% file: subsec_neuromuscular_models_cmaes.tex
\section{Review of Neuromuscular models}

\begin{wrapfigure}{r}{4.2cm}
\centering
\vspace{-15px}
\caption{\textbf{Neuromuscular Model.} }
\label{fig_nmm}
\includegraphics[width=0.25\textwidth]{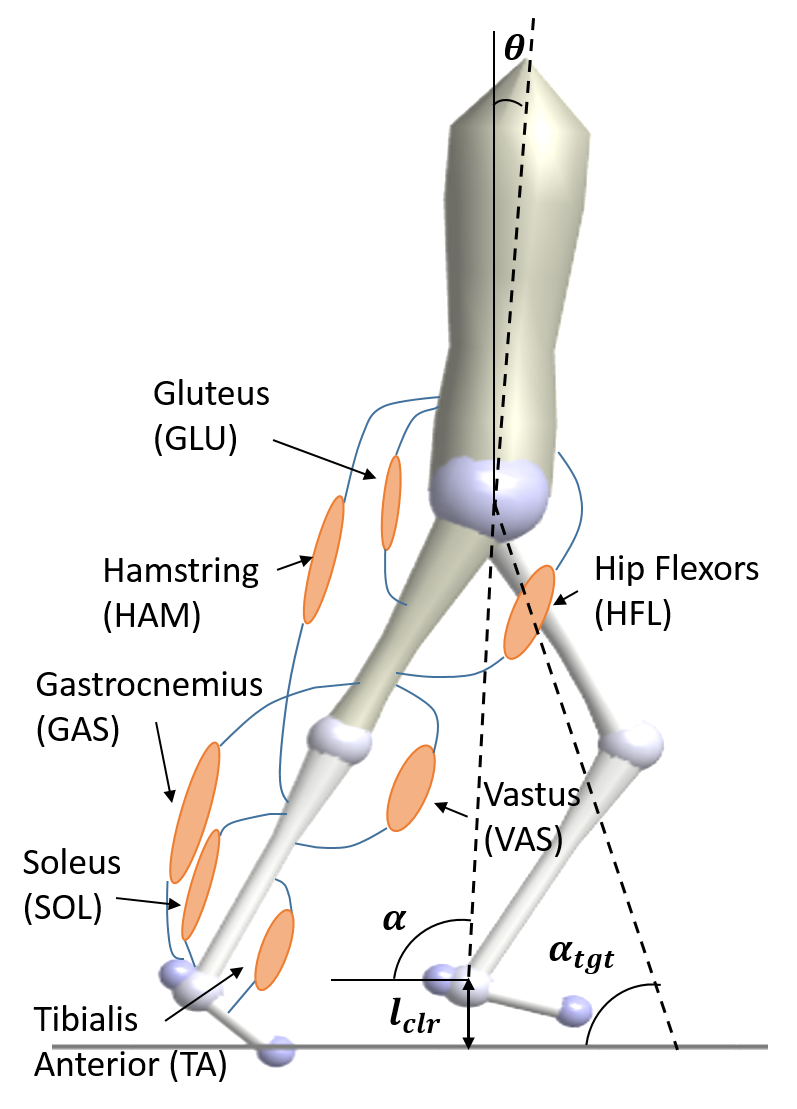}
\vspace{-25px}
\end{wrapfigure}

We use neuromuscular model policies, as introduced in \cite{geyer2010muscle}, as our controller for a 7-link planar human-like model. These policies use approximate models of muscle dynamics and human-inspired reflex pathways to generate joint torques, producing gaits that are similar to human walking in stance. \cite{desai} designed reflex laws for swing that enabled target foot-placement and leg clearance, by analyzing the double pendulum dynamics of the human leg. Integrating this swing control with the previous reflex control enables the model to overcome disturbances in the range of up to $\pm 10$ cm \cite{song2015neural}. 

\subsubsection{Neuromuscular Stance Control}   
In stance, each leg is actuated by 7 Hill-type muscles \cite{morrison1970mechanics}, consisting of the soleus (SOL), gastrocnemius (GAS), vastus (VAS), hamstring (HAM), tibialis anterior (TA), hip flexors (HFL) and gluteus (GLU), illustrated in Figure \ref{fig_nmm}. Together, these muscles produce torques about the hip, knee and ankle. The muscle force $F$ is a non-linear function of the muscle state $s^m$ and stimulus $S^m$, which when multiplied by the moment arm $r(\theta_i)$ gives the resultant torque on joint $i$:
\begin{equation*}
\tau_i^m = F(S^m, s^m)r(\theta_i),
\end{equation*}   
where $\tau_i^m$ is the torque applied by muscle $m$ on joint $i$ and $\theta_i$ is the joint angle.
 
Most of the muscle reflexes in stance are positive length or force feedbacks on the muscle stimulus. In general, the stimulus $S^m(t)$ for muscle $m$ is a function of the time delayed length or force signal $P^m$ times a feedback gain $K^m$:
\begin{equation*}
S^m(t) =  S_0^m + K^m \cdot P^m(t - \Delta t),
\end{equation*}
where $S_0^m$ is the pre-stimulus, $K^m$ is the feedback gain and $P^m$ is the time-delayed feedback signal of length or force. Some muscles can be co-activated and have multiple feedback signals from more than one muscle. The feedback gains $K^m$ described above are a subset of the parameters that we aim to tune in our optimization. The details of these feedback pathways can be found in \cite{song2015neural}.

This feedback structure generates compliant leg behaviour and prevents the knee from overextending in stance. To balance the trunk, feedback on the torso angle is added to the GLU stimulus:
\begin{equation*}
S^{GLU}_{torso}(t) = K_p^{stance}(\theta_{des} - \theta) -K_d^{stance}\dot{\theta},
\end{equation*}
where $K_p^{stance}$ is the position gain on the torso angle $\theta$ and $\theta_{des}$ is the desired angle. $K_d^{stance}$ is the velocity gain and $\dot{\theta}$ is the angular velocity.
Specifically, here are the stance parameters we optimize over, and their roles in the neuromuscular model:
 
\begin{enumerate}
\item $K^{GAS}$ : Positive force feedback gain on GAS
\item $K^{GLU}$ : Positive force feedback gain on GLU
\item $K^{HAM}$ : Positive force feedback gain on HAM
\item $K^{SOL}$ : Positive force feedback gain on SOL
\item $K^{TA}_{SOL}$ : Negative force feedback from SOL on TA
\item $K^{TA}$ : Positive length feedback on TA
\item $K^{VAS}$ : Positive force feedback on VAS
\item $K^{stance}_p$ : Position gain on feedback on torso angle
\item $K^{stance}_d$ : Velocity gain on feedback on torso velocity
\item $K_{mix}^{GLU}$ : Gain for mixing force feedback and feedback on angle for GLU
\end{enumerate}

\subsubsection{Swing Leg Placement Control}

The swing control is controlled by three main components -- target leg angle, leg clearance and hip control. Target leg angle is a direct result of the foot placement strategy which is a function of the velocity of the center of mass (CoM) $v$, and the as distance between the stance leg the CoM, and presented in \cite{simbicon}:
\begin{equation*}
\alpha_{tgt} = \alpha_0 + C_d d + C_v v,
\end{equation*}
where $\alpha_{tgt}$ is the target leg angle, $\alpha_0$ is the nominal leg angle, $\alpha_0$, $C_d$ and $C_v$ are parameters optimized by our control.

Leg clearance is a function of the desired leg retraction during swing. The knee is actively flexed until the leg reaches the desired leg clearance height, $l_{clr}$ and then held at this height, until the leg reaches a threshold leg angle. At this point, the knee is extended and allowed to reach the target leg angle $\alpha_{tgt}$. Details of this control can be found in \cite{desai}. As was noted in \cite{song2015neural}, and observed in our experiments, the control is relatively insensitive to the individual gains of the set-up in swing. It is sufficient to control the higher level parameters such as the desired leg clearance and target leg angle.

The third part of the control involves maintaining the desired leg angle $\alpha_{tgt}$ by applying a hip torque $\tau_{hip}^{\alpha}$:
\begin{equation*}
\tau_{hip}^{\alpha} = K_p^{swing}(\alpha_{tgt} - \alpha) - K_d^{swing}(\dot{\alpha}),
\end{equation*}
where $K_p^{swing}$ is the position gain on the leg angle, $K_d^{swing}$ is the velocity gain, $\alpha$ is the leg angle and $\dot{\alpha}$ is the leg angular velocity (see Figure \ref{fig_nmm}). 

More concisely, the swing parameters that we focus on in our optimization are the following:
\begin{enumerate}
\item $K_p^{swing}$ : Position gain on feedback on leg angle
\item $K_d^{swing}$ : Velocity gain on feedback on leg velocity
\item $\alpha_0$ : Nominal leg angle
\item $C_d$ : Gain on the horizontal distance between the stance foot and CoM
\item $C_v$ : Gain on the horizontal velocity of the CoM
\item $l_{clr}$ : Desired leg clearance
\end{enumerate}

Though originally developed for explaining human neural control pathways, these controllers have recently been applied to robots and prosthetics, for example in \cite{thatte} and \cite{van2015biped}. As demonstrated in \cite{song2015neural}, these models are indeed capable of generating a variety of locomotion behaviours for a humanoid model - for example, walking on flat, rough ground, turning, running, walking upstairs and on ramps. However, a full study of using these models to control biped robots still needs to be done. Whether these models will transfer well to robots with significantly different dynamics and inertial properties than humans needs to be explored. It is difficult to transfer these models to robots because of a large number of interdependent gains that need to be tuned. Typically, this is done using Covariance Matrix Adaptation Evolutionary Strategy \mbox{(CMA-ES)} \cite{hansen2006cma}, an evolutionary algorithm for difficult non-linear non-convex black-box optimization problems. Even though CMA-ES is useful for optimizing non-convex problems in high dimensions, it is not sample efficient and depends on the initial starting point. An optimization for 16 neuromuscular parameters takes 400 generations, around a day on a standard i7 processor and about 5,000 trials, as reported in \cite{song2015neural}. 

The large number of trials make it impossible to implement CMA-ES on a real robot. This is a shortcoming because often we find that after training the policies in simulation, they do not transfer well to the real robot, due to differences between simulation and real hardware.

%% file: subsec_bo_kernel.tex
\subsection{Kernels for Sequential Decision Making}

As described in section~\ref{subsec_bo_overview}, the kernel $k(\pmb{x}_i, \pmb{x}_j)$ captures how similar the cost function $f$ is expected to be for parameter vectors $\pmb{x}_i$ and $\pmb{x}_j$, and in the case of using Squared Exponential kernel, the similarity is a function of the Euclidean distance between $\pmb{x}_i, \pmb{x}_j \in \mathbb{R}^d$.

A more informed alternative is to use a kernel that specifically leverages the structure of the problem at hand, for example the resulting trajectories or behavior. Intuitively, a kernel that can better encode similarity among policies will be more sample-efficient, since it will be better able to generalize across policies with similar performance. The Behavior-Based Kernel (BBK) of \cite{wilson2014using} is one kernel that leverages structure in the trajectories generated by the evaluated policies.
To determine the similarity between policies given by vectors $\pmb{x}_i, \pmb{x}_j$, BBK uses the similarity of the trajectory distributions induced by the corresponding policies (instead of the default approach of utilizing Euclidean distance $|\pmb{x}_i - \pmb{x}_j|^2$).
While this could help in settings where computing a trajectory for a set of policy parameters is inexpensive, in the setting of robotic locomotion this requires running a simulation or executing the policy on the real hardware. This amounts to being as expensive as a cost function evaluation which makes BBK infeasible for such problems. Nonetheless, the idea of using auxiliary information in the kernel is promising, if this information can be pre-computed for a large portion of the policy space and made available during online optimization. We describe our approach for constructing such a kernel in the next section. Our kernel effectively incorporates domain knowledge available in bipedal locomotion and eliminates the need for computing full trajectories during online optimization. Instead, it uses behavior information from only a short part of the trajectories pre-generated during an offline phase.

%% file: subsec_new_distance_metric.tex
\subsection{Determinants of Gait Kernel}
\label{subsection_dog_metric}

Bipedal walking can be characterized with some basic metrics, called gait determinants, as described in \cite{Saunders}. The six determinants of gait deal with the conservation of energy and maintaining forward momentum during human walking. For developing our kernel, we focussed on the knee flexion in swing, ankle movement and center of mass trajectory. 

To compute the gait determinants of a given set of parameters, we run a short simulation for 5 seconds (as compared to 100 seconds for a complete trial). Next, we compute the score of the parameters on the following metrics $M_{1-5}$:
\begin{enumerate}
\item Is the knee flexed in swing?
\begin{align}
M_1  =  (\theta^{thr}_{high} > \theta_{knee}^{swing}) \land  (\theta_{knee}^{swing} > \theta^{thr}_{low})
\end{align}
Here, $\theta_{knee}^{swing}$ is the knee joint angle in swing, $\theta^{thr}_{high}$ and $ \theta^{thr}_{low}$ are the high and low thresholds on knee angle, similar to human data as described in \cite{winter}.
\item Is there heel-strike and toe-off?
\begin{align}
M_2 = (\theta_{ankle}^{strike} < 0) \land (\theta_{ankle}^{t.o.} > 0)
\end{align}
$\theta_{ankle}^{strike}$ is the ankle joint angle at heel-strike (start of stance) and $\theta_{ankle}^{t.o.}$ is the ankle joint angle at take-off (end of stance). The two conditions ensure heel-strike and toe-off respectively.
\item Is the center of mass movement approximately oscillatory between steps?
\begin{align}
M_3 = (Y_{CoM}^{strike} < Y_{CoM}^{midst}) \land
(Y_{CoM}^{t.o.} < Y_{CoM}^{midst})
\end{align}
$Y_{CoM}^{strike}$, $Y_{CoM}^{midst}$ and $Y_{CoM}^{t.o.}$ are center of mass heights at heel strike, mid stance and take-off. 
\item Is the torso leaning forward?
\begin{equation}
M_4 = \theta_{torso}^{mean} > 0
\end{equation}
$\theta_{torso}^{mean}$ is the mean torso angle which should be leaning forward for a energy efficient forward movement.
\item Deviation from average human walking speed
\begin{equation}
M_5 = ||v_{avg} - v_{human}||
\end{equation}
$v_{avg}$ and $v_{human}$ are the average simulator speed and the average human walking speed, $1.3m/s$ \cite{winter}.
\end{enumerate}

$M_{1-4}$ are binary $\in \{0,1\}$ and $M_5$ is continuous per step. These are then summed up to form the total score for \mbox{step $i$}:
\begin{equation}
score^{i} = \sum_{j = 1}^5 M_j^{i}
\end{equation}
The final metric is then computed as a sum of scores over all the steps:
\begin{equation}
\phi(\pmb{x}) = \sum_{i = 1}^{N} score^{i}
\end{equation}
where $N$ is the total number of steps in the first 5 seconds of simulation.

With this, a 16D point $\pmb{x}$ in the original parameter space now corresponds to a 1D point $\phi(\pmb{x})$ in this new feature space and we obtain our Determinants of Gait kernel:
\begin{equation*}
k(\pmb{x}_i, \pmb{x}_j) \rightarrow k(\phi(\pmb{x}_i), \phi(\pmb{x}_j))
\end{equation*}

$\phi(\pmb{x})$ is a very coarse measurement of the chances of the policy induced by $\pmb{x}$ resulting in stable walking movements over longer simulation periods. More importantly, points that lead to obviously unstable movements obtain a similar score of near zero, and are therefore grouped together. This kernel has no explicit information of the specific cost we are trying to optimize. It can very easily be used across multiple costs for walking behaviours, over slightly disturbed models as well as across multiple optimization methods.

%% file: subsec_experimental_setup.tex
In this section, we describe our experiments with the DoG kernel on two cost functions. First, we introduce our experimental settings and then move on to the results.

\subsection{Details of Experimental Setup: Cost Function and Algorithms Compared}
\label{subsection_experimental_setup}

To ensure that our approach can perform well across various cost functions, we conduct experiments on two different costs, constructed such that parameter sets achieving low cost also achieve stable and robust walking gaits. The first cost function varies smoothly over the parameter space:
\begin{align}
cost = \frac{1}{1+t} + \frac{0.3}{1+d} + 0.01 (s-s_{tgt}),
\end{align}
where $t$ is seconds walked, $d$ is the final hip position, $s$ is mean speed and $s_{tgt}$ is the desired walking speed (from human data). This cost encourages walking further and for longer through the first two terms, and penalizes deviating from the target speed with the last.

The second cost function is a slightly modified version of the cost used in~\cite{song2015neural} for experiments with \mbox{CMA-ES}. It penalizes policies that lead to falls in a non-smooth manner:
\begin{align}
cost_{CMA} = 		
    \begin{cases}
		300 - x_{fall} , \text{\small{if fall}} \\
		100 ||v_{avg} - v_{tgt}|| + c_{tr}, \text{\small{if walk}}\\
	\end{cases}
\end{align}
Here $x_{fall}$ is the distance travelled before falling, $v_{avg}$ is the average speed in simulation, $v_{tgt}$ is the target speed and $c_{tr}$ is the cost of transport. The first term directly penalizes policies that result in a fall, inversely to the distance walked. If the model walks for the simulation time, the cost is lower, ensured by the constants, and encourages policies that result in lower cost of transport and walk at target velocity. Since we have the same set of gains for left and right legs, the steadiness cost of the original cost~\cite{song2015neural} was unimportant.

In the following sections we compare the performance of several baseline and state-of-the-art optimization algorithms in simulation. Motivated by the discussion in~\cite{calandra2016bayesian}, we include the baseline of uniform random search. While this search is uninformed and not sample-efficient, it could (perhaps surprisingly) serve as a competitive baseline in non-convex high-dimensional optimization problems. 
We also provide comparisons with \mbox{CMA-ES}~\cite{hansen2006cma} and Bayesian Optimization with a Euclidean kernel (basic BO). Since we were optimizing a non-convex function in a 16D space, it was not feasible to calculate the global minimum exactly. To estimate the global minimum for the costs we used, we ran CMA-ES (until convergence) and BO with our domain kernel (for 100 trials) for 50 runs without model disturbances on flat ground. 
When reporting results, we plot the best results found in this easier setting as the estimated optimum for comparison.

For experiments with Bayesian Optimization we explored using two libraries: MOE developed by Yelp~\cite{MOE} and Matlab implementation from~\cite{gardner2014bayesian}.

%% file: subsec_model_disturbances.tex
\subsection{Model Disturbances}
\label{subsection_model_disturbances}

Most real robots have poor dynamic models, as well as unmodeled disturbances, like friction, non-rigid dynamics, etc which make simulations a poor representation of the real robot. There has been a lot of work done in identifying dynamic models of robots reliably, for example in \cite{atkeson}. However, while such methods can definitely help bring simulators close to the real robot, there are other discrepancies like non-rigid dynamics and  friction which are still very hard to model. As a result, often controllers that work well in simulation lead to poor performance on the real robot. In such cases, ideally, we would like to have optimization techniques that quickly adapt to this slightly different setting and find a new solution in a few cost function evaluations. 

To test if our approach is capable of generalizing to unforeseen disturbances, modeling and environmental perturbations, we conduct our experiments on models with mass and inertial disturbances and on different ground profiles. We perturb the mass of each link, inertia and center of mass location randomly by up to $15\%$ of the original value. For mass/inertia we randomly pick a variable from a uniform distribution between $[-0.15,\ 0.15] \cdot M$, where $M$ is the original mass/inertia of the segment. Similarly we change the location of the center of mass by $[-0.15,\ 0.15] \cdot L/2$, where $L$ is the length of the link. These disturbances are different for each run of our algorithm, hence we test a wide range of possible modelling disturbances.
For the ground profiles, we generate random ground height disturbances of upto $\pm 8cm$ per step.

%% file: subsec_experiments_kernel.tex
\subsection{Experiments with DoG Kernel}

We pre-compute Determinants of Gait (DoG) kernel scores for 100,000 parameter sets, which takes 7-10 hours on a modern desktop to speed up our computations. These samples are generated using a Sobol sequence~\cite{bratley1988algorithm} on an undisturbed model on flat ground. Thereafter, the same kernel is used for all the experiments described below.

In experiments with Bayesian Optimization, we were directly able to replace the Euclidean distances of a squared exponential kernel with the distance in the DoG kernel space:
\vspace{-7px}
\begin{equation}
k_{DoG}(\pmb{x}_i, \pmb{x}_j) = \exp\Big(-\frac{1}{2} \|\phi(\pmb{x}_i) - \phi(\pmb{x}_j)\|^2 \Big)
\vspace{-3px}
\end{equation}
with $\phi(\pmb{x}_i)$ as described in Section~\ref{subsection_dog_metric}. We used a Matlab implementation of Bayesian Optimization from~\cite{gardner2014bayesian} and used a pre-sampled grid when considering next candidates for optimization. This allowed us to reuse our pre-computed scores to speed up kernel computations, but restricted us to only use these pre-sampled parameter sets. This can be harmful if an optimal set was not sampled; we sampled a dense grid to decrease the probability of this happening. 

\begin{figure}[t]
\centering
\caption{Top row: a policy that generates successful walking on flat ground could fail on rough ground. Bottom row: optimization on rough ground finds policies that walk, even though pre-computation for DoG kernel is done using unperturbed model on flat ground.}
\label{fig_bo_locomotion_visualization_flat}
\vspace{-7px}
\includegraphics[width=0.20\textwidth]{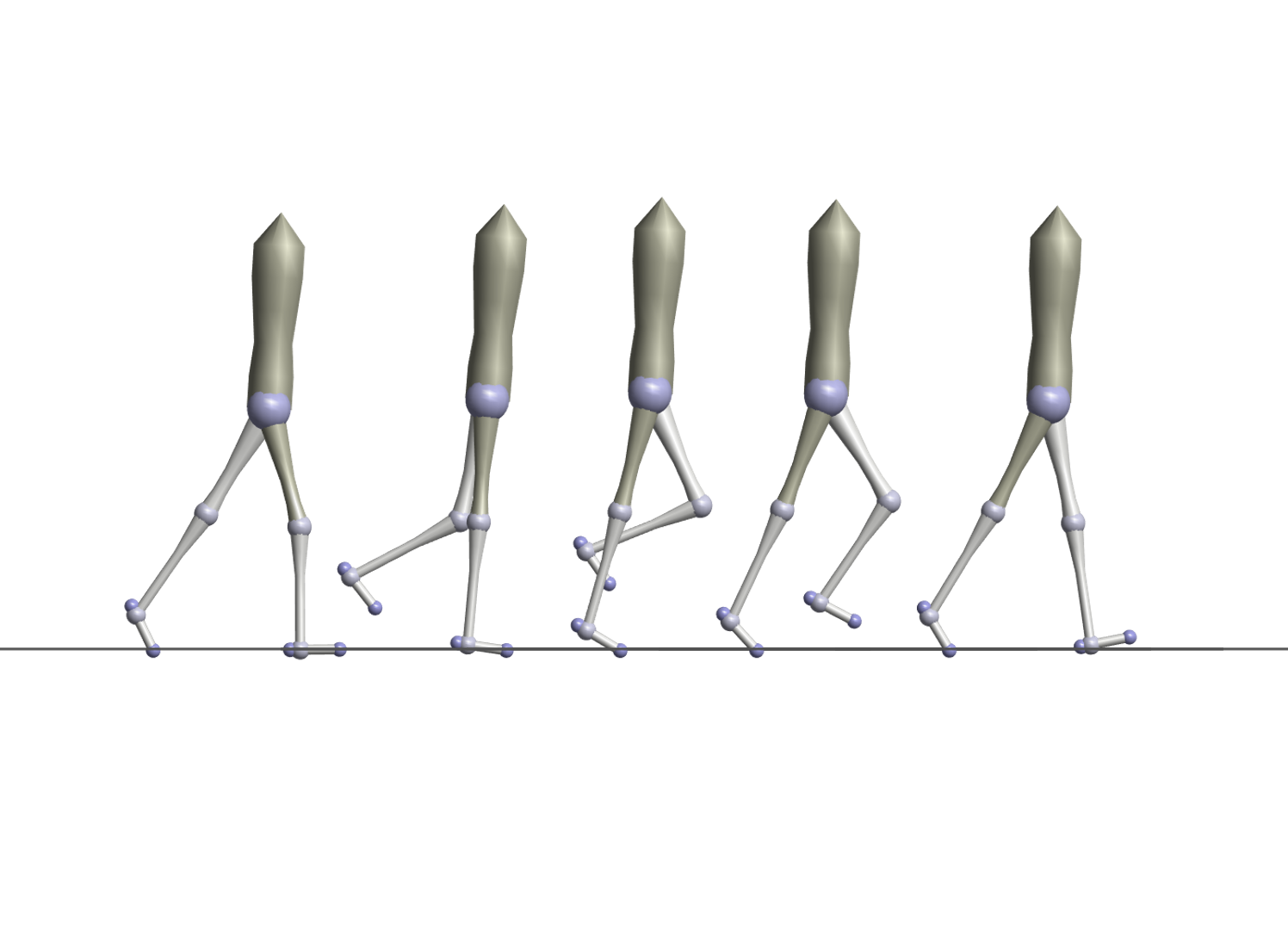}
\hspace{10px}
\vspace{-20px}
\includegraphics[width=0.25\textwidth]{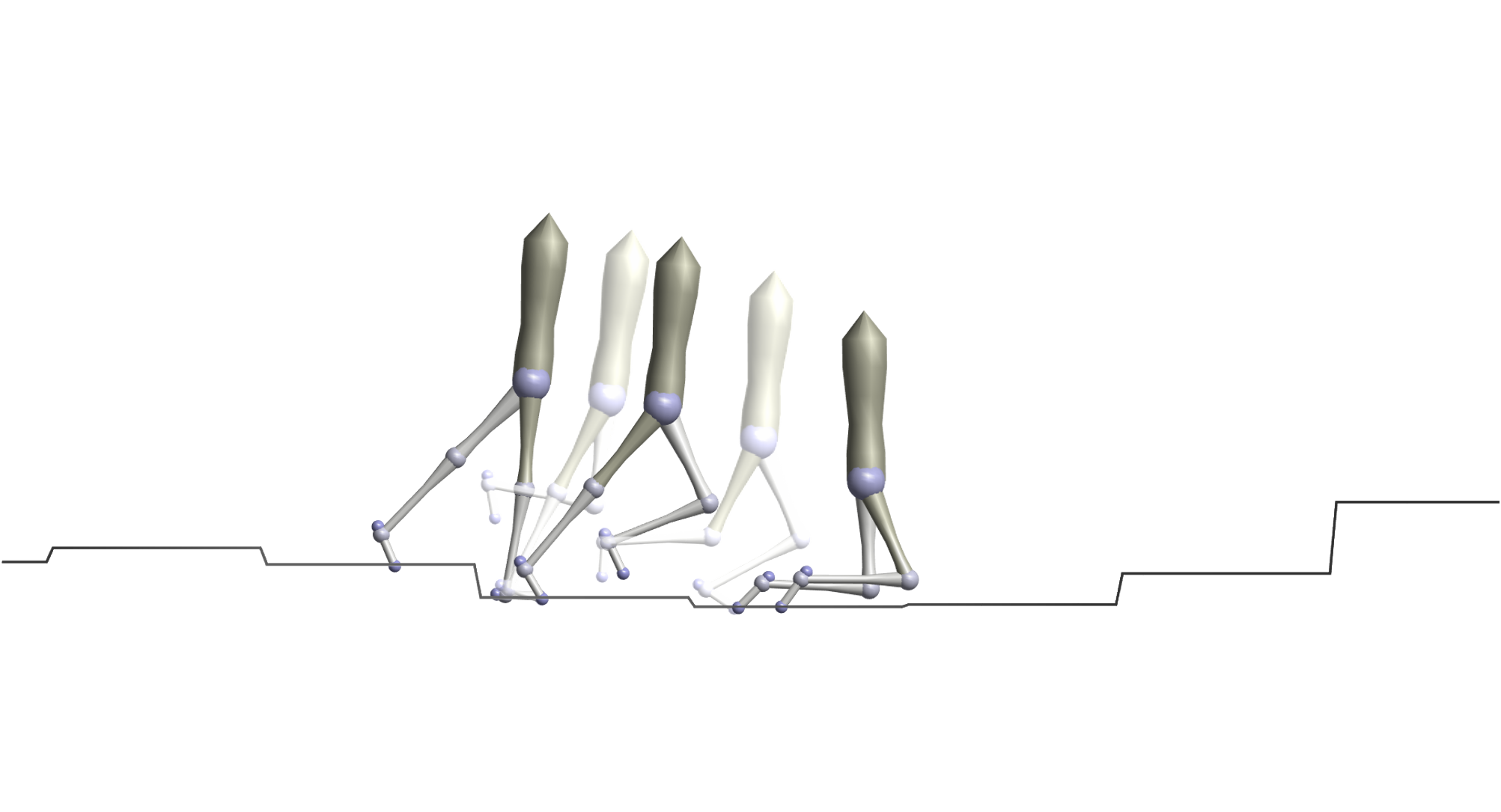}
\label{fig_bo_locomotion_visualization_rough}
\includegraphics[width=0.3\textwidth]{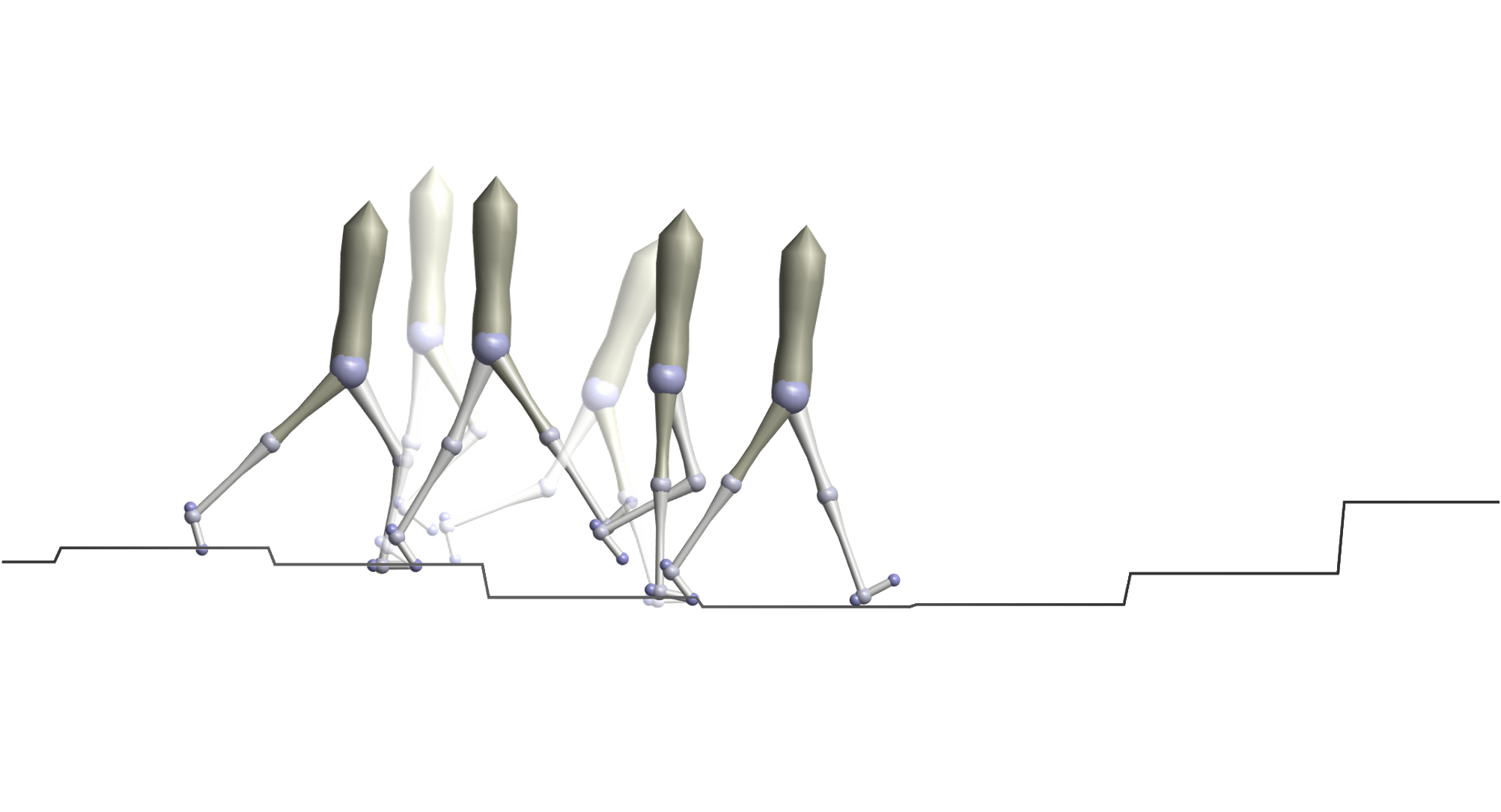}
\vspace{-25px}
\end{figure}

Our DoG scores were obtained from an unperturbed model of our system on flat ground. Our experimental results, however, were obtained on settings with different ground profiles and model disturbances (as discussed in \mbox{Section~\ref{subsection_model_disturbances}}). 
These perturbed settings were designed such that originally optimal set of policy parameters would likely become sub-optimal. This is illustrated in the top and middle rows of Figure~\ref{fig_bo_locomotion_visualization_flat}, where the policy performing well on flat ground falls on rough ground. 
This shows that our perturbations were indeed significant. After using the kernel for the optimization in these perturbed settings, we observed that best policies found were able to walk on rough ground (the lower part of Figure~\ref{fig_bo_locomotion_visualization_flat}). 
This suggests that our kernel can be used to find optimum in settings significantly different than those it was created on.

All the experiments described below are done for 50 independent runs, each with a unique set of modeling disturbances and a different ground profile for rough ground walking. Each run  consists of 100 trials or cost function evaluations, in which the optimization algorithm evaluates a parameter set for 100 seconds of simulation. Note that the disturbances and ground profiles remain constant across each run (and 100 trials).

\subsubsection{Experiments on the smooth cost function}

\begin{figure}[t]
\centering
\caption{Experiments using DoG kernel on rough ground with model disturbances on the smooth cost over 50 runs. Basic BO line (green triangles) was obtained using BO with a Euclidean distance kernel. Estimated optimum (yellow line) was obtained as described at the end of Section~\ref{subsection_experimental_setup}. BO with DoG kernel (blue pentagons) used the kernel we described in Section~\ref{subsection_dog_metric} and was substantially more sample-efficient than the alternative approaches.}
\vspace{0.25cm}
\label{fig_bo_locomotion}
\includegraphics[width=0.42\textwidth]{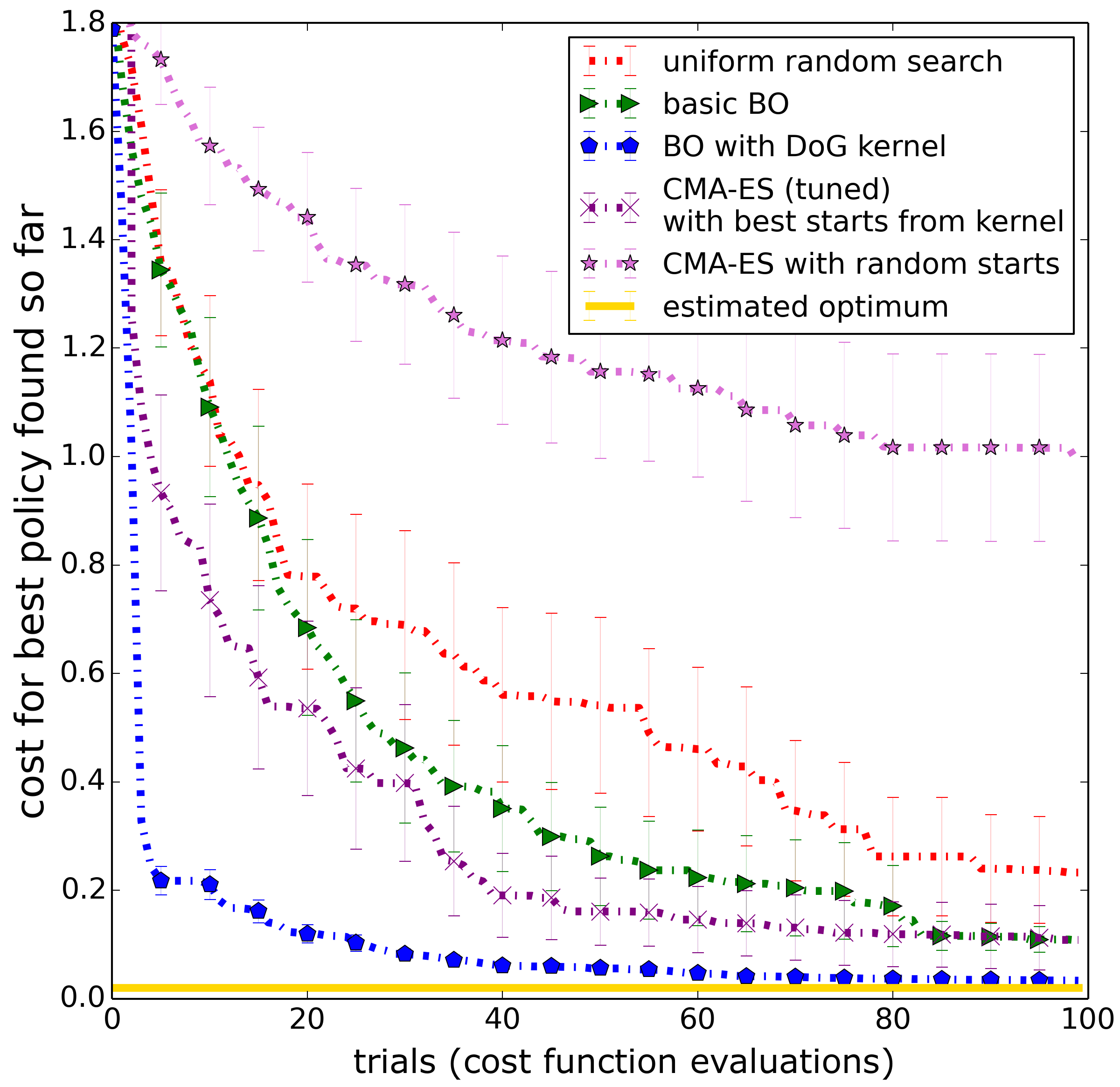}
\vspace{-10px}
\end{figure}

Figure~\ref{fig_bo_locomotion} shows results of our experiments using the DoG kernel on the smooth cost.
For BO with DoG kernel, 25-30 cost function evaluations were sufficient to find points that corresponded to robot model walking on a randomly generated rough ground with $\pm 8 cm$ disturbance. This is in contrast to basic BO that did not find such results in under 100 trials. 

To let CMA-ES also benefit from the kernel, we started each run from one of the best 100 points for the DoG kernel. After tuning the $\sigma$ parameter of CMA-ES to make it exploit more around the starting point, we were able to find policies that resulted in walking on rough ground after 65-70 cost function evaluations on most runs. On the other hand, \mbox{CMA-ES} starting from a random initial point was not able to find walking policies in 100 evaluations.

These results suggest that DoG scores successfully captured useful information about the parameter space and were able to effectively focus BO and CMA-ES on the promising regions of the policy search space. 

\subsubsection{Experiments with the non-smooth cost}

\begin{figure}[t]
\centering
\caption{Experiments using DoG kernel on rough ground with model disturbances on the non-smooth cost over 50 runs. Policies with costs below 100 generate walking behaviour for 100 seconds in simulation. None of the optimization methods find optimal policies in all the runs and hence the mean cost is higher than the estimated optimum.}
\vspace{0.25cm}
\label{fig_bo_locomotion_non_smooth}
\includegraphics[width=0.45\textwidth]{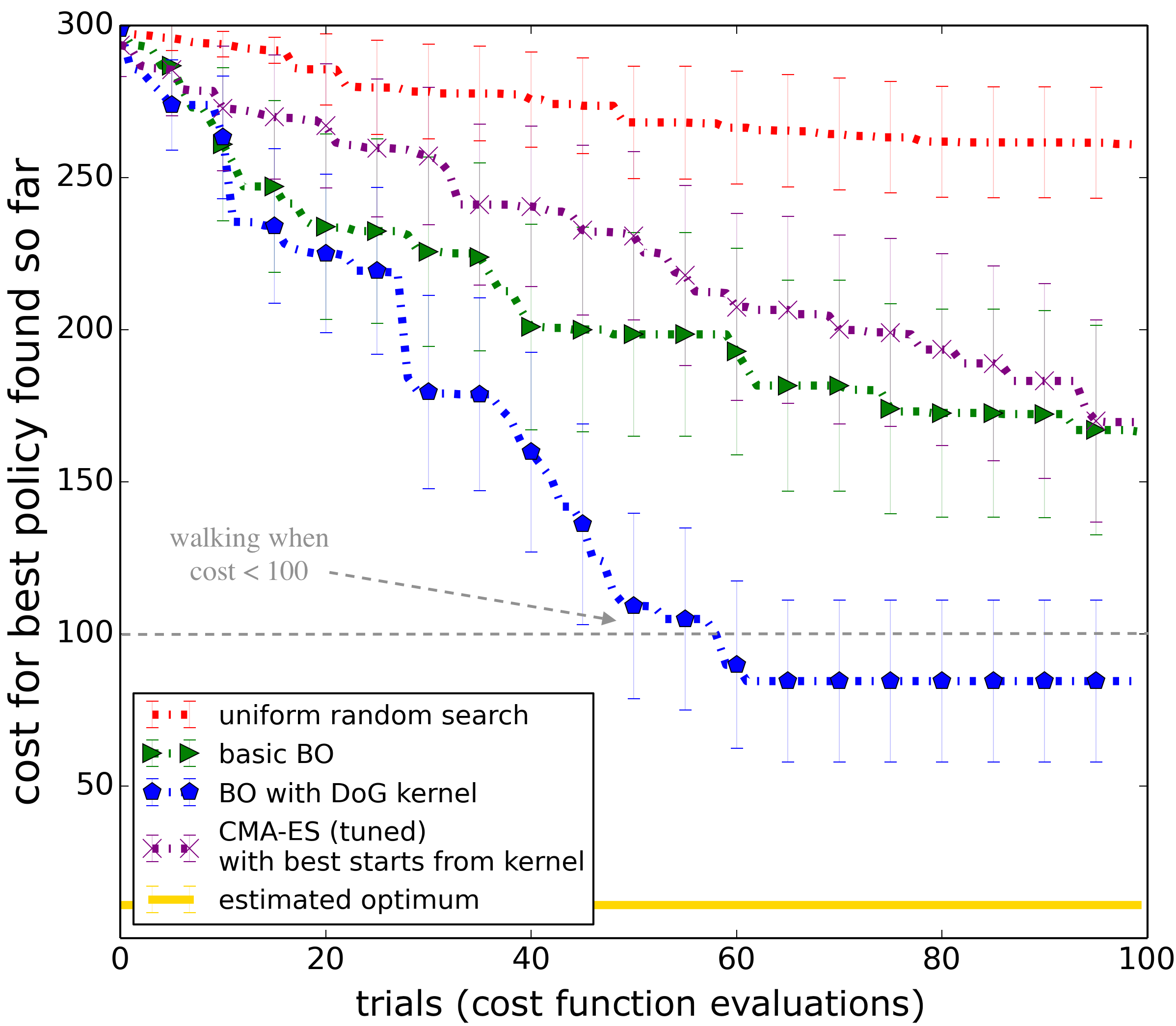}
\vspace{-10px}
\end{figure}

We observed good performance on the non-smooth cost function (Figure \ref{fig_bo_locomotion_non_smooth}), though it was not as remarkable as the smooth cost. BO with kernel still outperformed all other methods by a margin, but this different cost seems to hurt BO and \mbox{CMA-ES} alike. Since this cost is discontinuous, there is a huge discrepancy between costs for parameters that walk and those that don't. If no walking policies are sampled, BO learns little about the domain and samples randomly, which makes it difficult to find good parameters. Hence not all runs find a walking solution. BO was able to find successful walking in $74\%$ of cases on rough ground with $\pm 6 cm$ disturbance in less than 60 trials/evaluations. \mbox{CMA-ES} starting from a good kernel point was able to do it in $40\%$ of runs. 

This showed that our kernel was indeed independent of the cost function to an extent, and worked well on two very different costs. We believe that the slightly worse performance on the second cost is because of the cost structure, rather than a kernel limitation, as it still finds walking solutions for a significant portion of runs.

\subsubsection{Experiments on different terrains}

\begin{figure}[t]
\centering
\caption{Optimized policies walking up and down a $12.5 \%$ ramp.}
\label{fig_ramp}
\vspace{3px}
\includegraphics[width=0.4\textwidth]{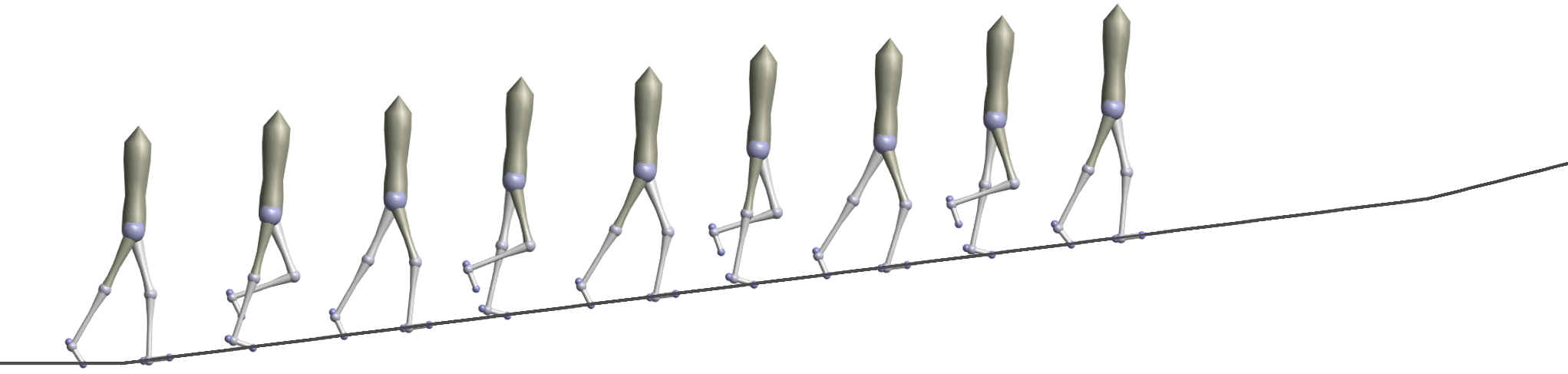}
\includegraphics[width=0.4\textwidth]{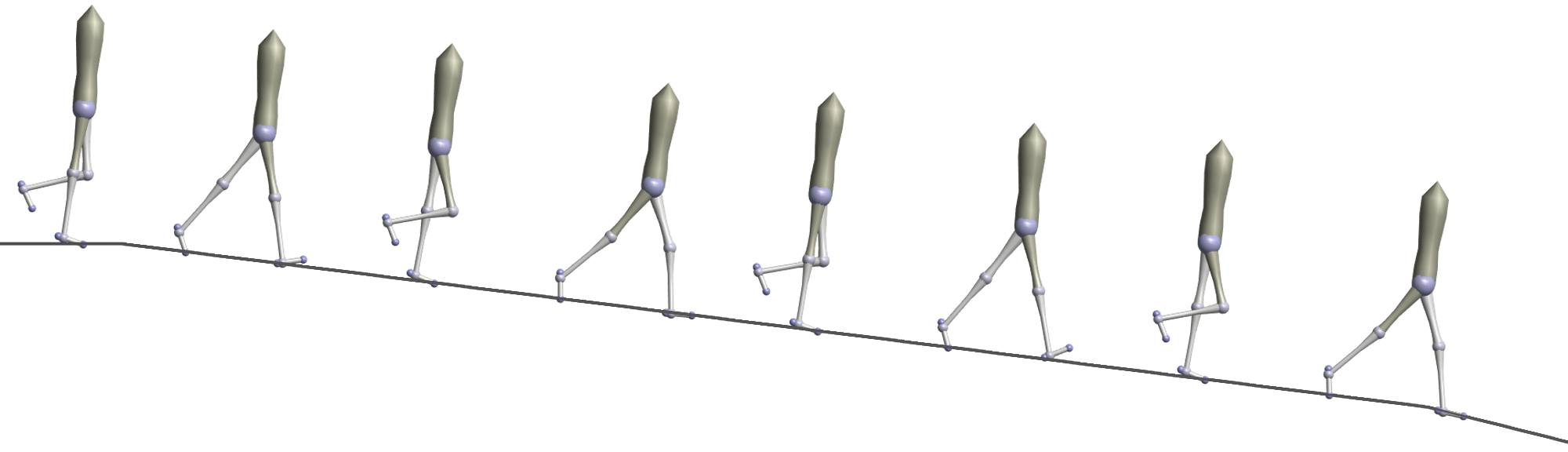}
\end{figure}

We also optimized on ramps -- sloping upwards, as well as downwards. The ramp up and down ground slopes were gradually increased every $20m$, until the maximum slope was reached. The maximum slopes for going down and going up were $20\%$ ($tan(\theta) = 0.2$). BO with DoG kernel was able to find parameters that walked for 100 seconds in $50\%$ of cases in ramp up and $90\%$ in ramp down. Example optimized policies walking up and down slope are shown in Figure \ref{fig_ramp}.

We believe the reason we could not find walking policies on ramps in all runs, was that we are not optimizing the hip lean, which was noted to be crucial for this profile in \cite{song2015neural}. Since we did not consider this variable when generating our 16 dimensional kernel, it was not trivial to optimize over it without re-generating the grid. Similarly, we found that we could not find any policies that climbed up stairs. Perhaps this could be achieved when optimizing over a much larger set of parameters, as in \cite{song2015neural}.

In the future, we would like to include more variables for optimizing over different terrains, and include them as part of the kernel. 

%% file: sec_conclusions.tex
In this work we focused on sample-efficiently finding walking policies for a bipedal neuromuscular model. This high dimensional optimization problem proved challenging for standard Bayesian Optimization. 
So, we introduced the Determinants of Gait (DoG) metric and constructed the corresponding DoG kernel to effectively incorporate domain knowledge into the kernel. For our experiments we pre-computed the kernel on flat ground with an unperturbed model, and then tested in more challenging settings. We demonstrated that our approach offers improved performance for learning walking patterns on different ground profiles, like rough ground, ramp up and ramp down, all with various unknown inertial disturbances to the original model. 

Our results motivated us to consider several directions of future work. One of the next steps would be to experiment with learning more parameters of the neuromuscular model. Adjusting more parameters would allow us to fine-tune walking behaviors for more challenging settings like stairs and steeper ramps. This would also make the problem more challenging because of the increase in the dimensionality of the search space. An informed kernel could provide robust performance by simplifying the search. We also would like to experiment with different ways of computing final DoG scores, perhaps by constructing a k-dimensional vector of individual metrics instead of collapsing them into a scalar score. And most importantly, we want to experiment with our approach on real hardware. We developed our experimental setup with future hardware experiments in mind, so we hope our approach would offer the needed sample efficiency to enable learning control policy parameters on the real hardware efficiently and adaptively.